\documentclass[letterpaper]{article} 
\usepackage{aaai2026}  
\usepackage{times}  
\usepackage{helvet}  
\usepackage{courier}  
\usepackage[hyphens]{url}  
\usepackage{graphicx} 
\usepackage{natbib}  
\usepackage{caption} 
\usepackage{algorithm}
\usepackage{algorithmic}

\usepackage{mathrsfs}
\usepackage{amsmath}
\usepackage{amssymb}
\usepackage{multirow}
\usepackage{booktabs}
\usepackage{makecell}
\usepackage{subcaption}

\usepackage{xcolor}

\newcommand{\set}[1]{\mathcal{#1}}

\usepackage{cuted}

\usepackage{newfloat}
\usepackage{listings}
\DeclareCaptionStyle{ruled}{labelfont=normalfont,labelsep=colon,strut=off} 
\floatstyle{ruled}
\newfloat{listing}{tb}{lst}{}
\floatname{listing}{Listing}
\title{Enhancing Generalization of Depth Estimation Foundation Model via Weakly-Supervised Adaptation with Regularization}
\author {
    Yan Huang\textsuperscript{\rm 1},
    Yongyi Su\textsuperscript{\rm 1},
    Xin Lin\textsuperscript{\rm 2},
    Le Zhang\textsuperscript{\rm 3},
    Xun Xu\textsuperscript{\rm 4}\thanks{Corresponding author.}
}
\affiliations {
    \textsuperscript{\rm 1}South China University of Technology\\
    \textsuperscript{\rm 2}Guangzhou University\\
    \textsuperscript{\rm 3}University of Electronic Science and Technology of China\\
    \textsuperscript{\rm 4}Institute for Infocomm Research (I\textsuperscript{2}R), A*STAR\\
    \{eeyhuang, eesuyongyi\}@mail.scut.edu.cn, linxin94@gzhu.edu.cn,\\
    zhangleuestc@gmail.com, alex.xun.xu@gmail.com
}

\usepackage{bibentry}

\begin{document}

\maketitle

\begin{abstract}
The emergence of foundation models has substantially advanced zero-shot generalization in monocular depth estimation (MDE), as exemplified by the Depth Anything series. However, given access to some data from downstream tasks, a natural question arises: can the performance of these models be further improved? To this end, we propose WeSTAR, a parameter-efficient framework that performs \textbf{We}akly supervised \textbf{S}elf-\textbf{T}raining \textbf{A}daptation with \textbf{R}egularization, designed to enhance the robustness of MDE foundation models in unseen and diverse domains. We first adopt a dense self-training objective as the primary source of structural self-supervision. To further improve robustness, we introduce semantically-aware hierarchical normalization, which exploits instance-level segmentation maps to perform more stable and multi-scale structural normalization. Beyond dense supervision, we introduce a cost-efficient weak supervision in the form of pairwise ordinal depth annotations to further guide the adaptation process, which enforces informative ordinal constraints to mitigate local topological errors. Finally, a weight regularization loss is employed to anchor the LoRA updates, ensuring training stability and preserving the model's generalizable knowledge. Extensive experiments on both realistic and corrupted out-of-distribution datasets under diverse and challenging scenarios demonstrate that WeSTAR consistently improves generalization and achieves state-of-the-art performance across a wide range of benchmarks.
\end{abstract}

\begin{figure}[t]
    \centering
    \includegraphics[width=\linewidth]{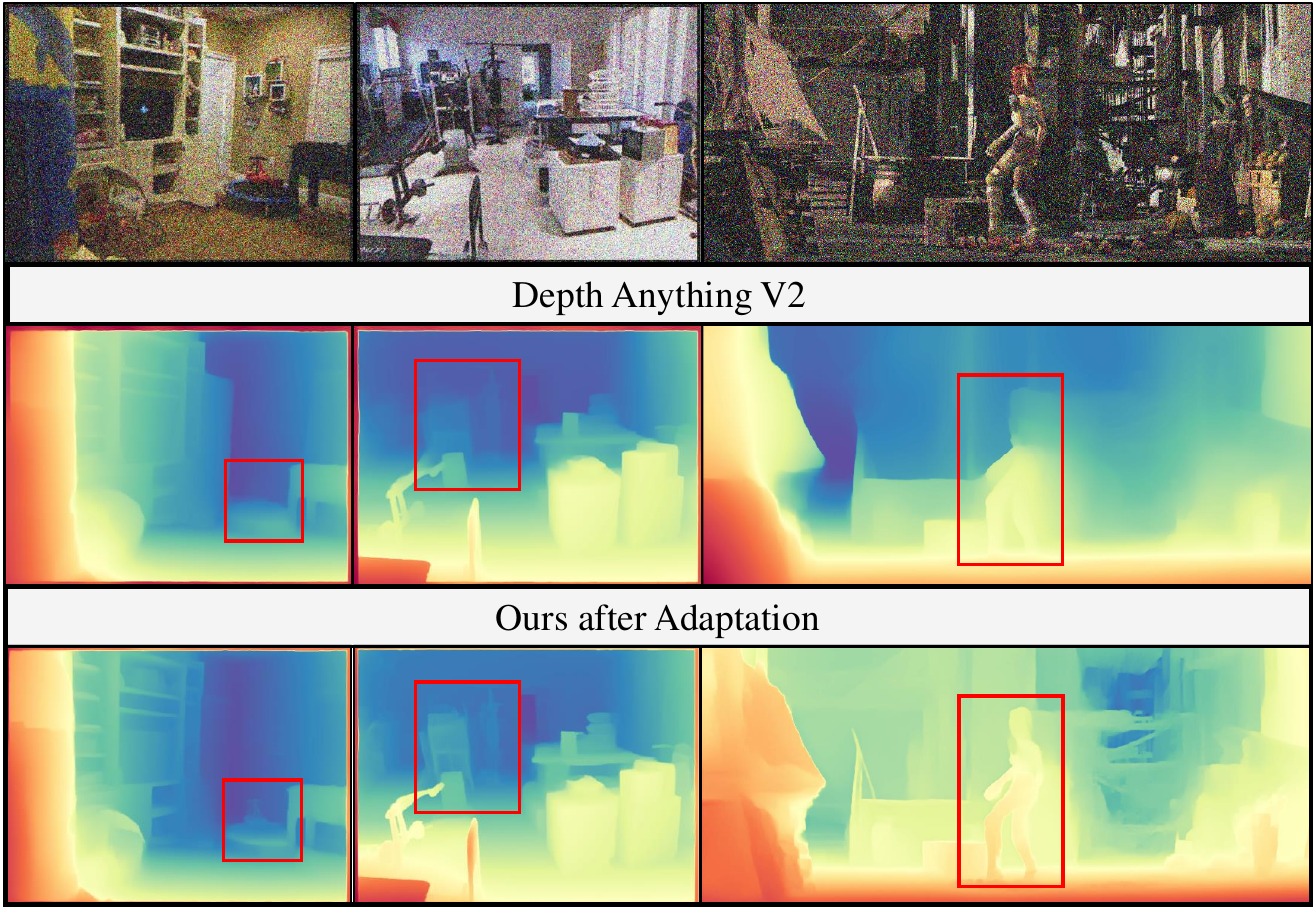}
    \caption{Illustration of MDE on unseen test samples with zero-shot results of model and model after adaptation.}
    \vspace{-0.5cm}
    \label{fig:teaser}
\end{figure}

\section{Introduction}

Monocular depth estimation is crucial for  applications such as stereo conversion~\cite{xie2016deep3d}, augmented reality~\cite{ganj2024mobile}, and 3D scene reconstruction~\cite{yin2022towards}. Recent progress has led to foundation models trained on large-scale combinations of labeled and unlabeled data, with models like those by ~\cite{yang2024depth, yang2024depthv2} showing strong generalization to diverse, unseen domains.

Despite these advancements, it remains unclear how much further performance can improve when data from downstream tasks becomes available. 
Figure~\ref{fig:teaser} shows that zero-shot predictions remain imperfect especially under distribution shifts. Performance can be further improved via adaptation.
In image segmentation, foundation models have been effectively adapted using unlabeled or weakly labeled task-specific data~\cite{zhang2024improving, zhao2025fishertune}, often leveraging self-training~\cite{sohn2020fixmatch}, self-supervised learning~\cite{chen2020simple, liu2021ttt++}, and distribution alignment~\cite{su2022revisiting}.

However, extending such techniques to depth estimation, essentially a regression task, poses unique challenges. Unlike classification, regression complicates self-training due to the difficulty of generating reliable pseudo-labels. Inaccurate labels can mislead the model during adaptation, especially in unseen domains. As shown in Figure~\ref{fig:sintel_diode_epochs}, naïve self-training can even degrade performance over time. 
{Moreover, the dense self-training (ST) objective often provides only marginal gains on benchmarks, mainly due to the strong geometric understanding of the pre-trained model,  which limits further improvement from ST alone. Finally, an overly aggressive adaptation process of self-training can be detrimental. It risks over-specializing the model to the specific domains, leading to overfitting and catastrophic forgetting, where the powerful, generalizable knowledge from the original pre-training is compromised.}

{Our framework addresses these limitations through a synergistic design. To mitigate confirmation bias and provide targeted, non-redundant supervision, we complement dense self-training with a weakly supervised objective~\cite{chen2016single}, incorporating a minimal set of relative depth annotations that correct fine-grained topological errors and introduce label signals independent of the model’s pseudo-labels.
In other words, the weakly supervised loss introduces sparse but strong ordinal constraints, while the dense self-training objective provides global structural alignment.
}

{To enhance adaptation efficiency and prevent catastrophic forgetting, motivated by robustness concerns in pre-trained models~\cite{zhao2023pitfalls}, we introduce additive low-rank adapters~\cite{hu2022lora} to constrain the model near the original. We further introduce a regularization that penalizes LoRA parameter updates, anchoring the model to the strong pre-trained depth prior and limiting the influence of noisy pseudo-labels and sparse ordinal constraints, suppressing the geometric distortions induced by weak supervision and mitigating the confirmation bias of self-training.
}

{We also observe that self-training for MDE must address scale and shift ambiguity between teacher pseudo-labels and student predictions. While hierarchical depth normalization (HDN)~\cite{zhang2022hierarchical, he2025distill} was designed to mitigate scale variation and depth discontinuities, its reliance on content-agnostic grids overlooks semantic context, often resulting in inferior performance. To address this, we propose using an external segmentation foundation model~\cite{ravisam2} to inject semantic information as masks, enabling semantic-aware HDN.}

We refer to our adaptation framework as \textbf{WeSTAR}.
WeSTAR enables fine-tuning of pre-trained models using minimal data from the target domain, resulting in improved performance on downstream tasks. We validate the method on both unseen image domains and corrupted inputs~\cite{kong2023robodepth}, simulating real-world out-of-distribution scenarios. This makes it especially suitable for depth estimation tasks in specific contexts where limited data with weak annotations can be collected for adaptation.

Our contributions are summarized as follows:

\begin{itemize} 
\item We address the challenge of generalizing depth estimation foundation models to unseen data distributions through a low-rank regularized self-training framework with semantic-aware hierarchical normalization. 
\item We incorporate low-cost weak labels, specifically pariwise ordinal depth annotations, to further enhance the robustness of the adaptation process. 
\item Our proposed method WeSTAR demonstrate robust generalization across diverse and challenging scenarios, including novel tasks and corrupted datasets. \end{itemize}

\section{Related Work}

\noindent \textbf{Monocular Depth Estimation:}
Monocular depth estimation (MDE) has advanced rapidly with deep learning, supported by architectural innovations and large-scale datasets~\cite{he2016deep, dosovitskiy2020image, geiger2013vision}. Early supervised~\cite{eigen2014depth, bhat2021adabins} and self-supervised methods~\cite{godard2017unsupervised, godard2019digging} achieved strong performance but struggled to generalize due to dataset biases. Transfer learning with pretrained backbones (e.g., ImageNet~\cite{deng2009imagenet}) improved generalization but still required task-specific tuning. Recent foundation models like MiDaS~\cite{ranftl2020towards} and Depth Anything~\cite{yang2024depth, yang2024depthv2} enhance robustness by training on large-scale data.

\noindent \textbf{Source-Free Domain Adaptation:}
Source-Free Domain Adaptation (SFDA)\cite{liang2020we} adapts models without access to source data. Our work aligns with SFDA, as zero-shot generalization often relies solely on pretrained models. In classification, early SFDA methods reconstruct source-like features\cite{ding2022source} or augment target samples~\cite{jing2022variational, hwang2024sfda}. More recent approaches use contrastive learning~\cite{zhang2022divide, chen2022contrastive}, pseudo-labeling~\cite{liang2020we}, and iterative self-training~\cite{karim2023c}, along with diversity constraints~\cite{mitsuzumi2024understanding} and feature alignment~\cite{lee2023feature, su2022revisiting}. For regression, TASFAR~\cite{he2024target} calibrates models using pseudo-label distributions. In depth estimation, some methods discretize depth into ordinal labels~\cite{yi2023test}, while others combine scale alignment and self-supervision~\cite{li2023test} to adapt without target supervision.

\noindent\textbf{Robustness in Depth Estimation:}
MDE models often fail under severe domain shifts like weather or noise, as shown in RoboDepth~\cite{kong2023robodepth}. Our analysis reveals that even state-of-the-art models (e.g., Depth Anything V2) degrade significantly under compounded corruptions. While this vulnerability is known, systematic methods to address it remain limited. We present a principled solution to enhance robustness under such shifts.

\noindent \textbf{Weakly Supervised Depth Estimation:}
Due to the difficulty of obtaining dense depth labels, alternatives like self-supervision from videos~\cite{godard2017unsupervised, zhao2022monovit} and weak supervision via ordinal or semantic cues~\cite{chen2016single} have gained popularity. Inspired by this, we introduce a structured pairwise ranking loss that enforces transitive ordinal relations, promoting a globally coherent depth manifold.

\begin{figure*}[!t]
    \centering
    \includegraphics[width=0.85\linewidth]{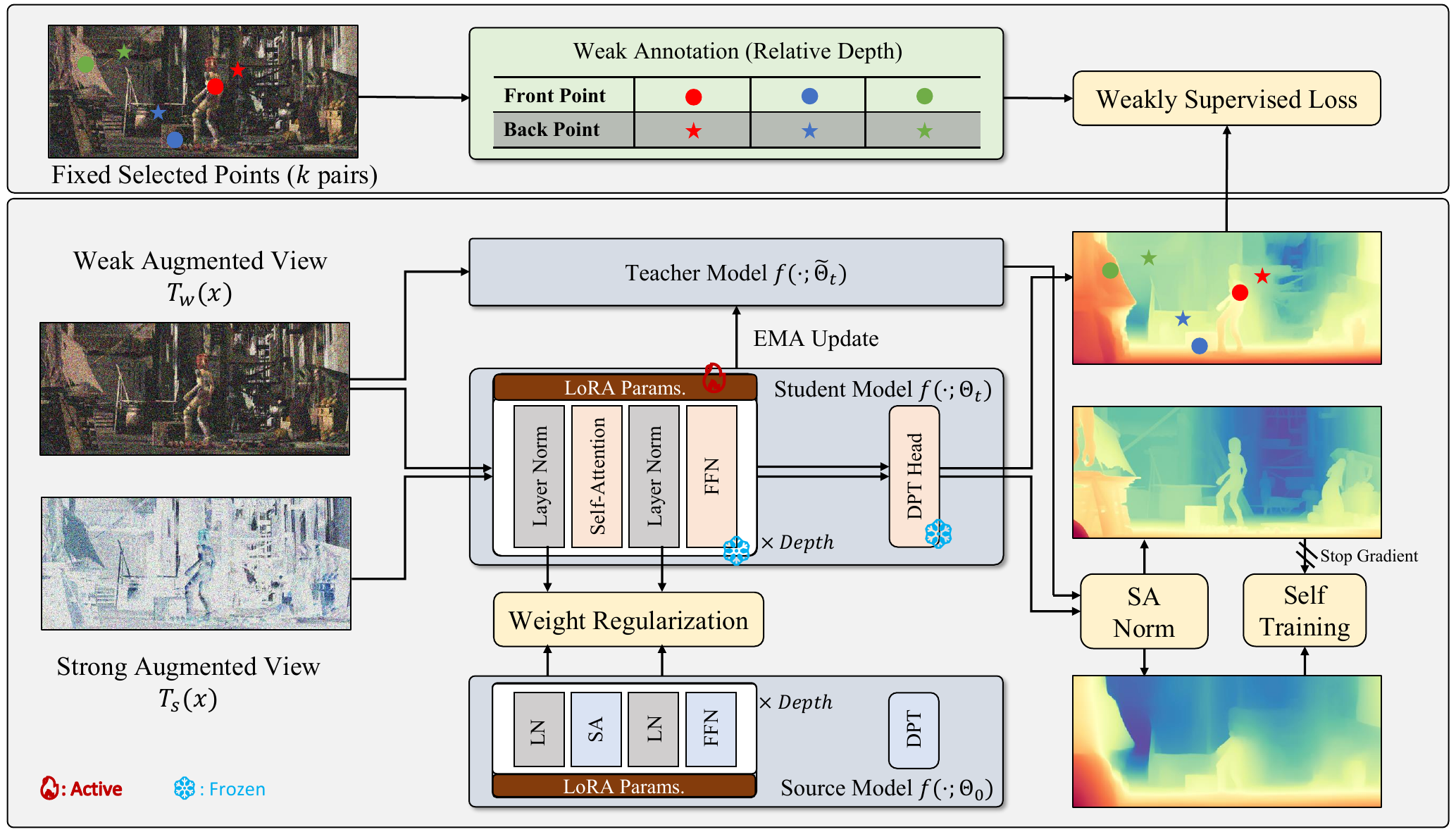}
    \caption{Illustration of the overall framework. Two augmentations are applied and the teacher model $\tilde{\Theta}$ generates pseudo labels. Self-training is regularized by model weights consistency and optionally weak labels to adapt pre-trained foundation model.}
    \vspace{-0.2cm}
    \label{fig:pipeline}
\end{figure*}

\section{Methodology}

\subsection{Problem Formulation}

We begin by formally defining the task of adapting a pre-trained monocular depth estimation foundation model to a target domain. Let $\set{D}_s = \{x_i, d_i\}$ denote a labeled source dataset, where $x_i$ represents input RGB images and $d_i$ their corresponding ground-truth depth maps. A depth estimation model $f(x; \Theta)$ is trained on this data using supervised or semi-supervised techniques~\cite{yang2024depth}.
Although the pre-trained model often generalizes well to many unseen domains, we hope to further enhance its effectiveness on the unseen domains. Our goal is to adapt $f(x; \Theta)$ to a new target domain dataset $\set{D}_{te} = \{x_j\}$ by utilizing a small set of either unlabeled or weakly labeled training samples $\set{D}_{tr} = \set{D}_{tr}^w \cup \set{D}_{tr}^u$, where $\set{D}_{tr}^w = \{x_j, w_j\}$ is weakly labeled and $\set{D}_{tr}^u = \{x_j\}$ is unlabeled. An overview of the framework is presented in Figure~\ref{fig:pipeline}.

\subsection{Domain Adaptation via Self-Training}
We first consider a fully unsupervised adaptation scenario, where $\set{D}_{tr}^w = \emptyset$.
Self-training (ST) is a well-established technique for unsupervised adaptation of pre-trained models, particularly successful in classification tasks~\cite{su2022revisiting, liang2020we}. ST typically adopts a teacher-student architecture~\cite{tarvainen2017mean}, where the teacher model generates pseudo labels to guide the student model's learning.

We employ an Exponential Moving Average (EMA) update rule for maintaining the teacher weights:  
$\tilde{\Theta}_{t+1} = \alpha \tilde{\Theta}_t + (1 - \alpha)\Theta_{t+1}$, where $\alpha$ is a smoothing factor.
Each unlabeled sample is augmented with weak ($T_w$) and strong ($T_s$) perturbations before being passed to the teacher and student models, respectively. The self-training loss is defined as:
\begin{equation}\label{eq:selftrain}
\begin{aligned}
    &\mathcal{L}_{st} = \frac{1}{|\set{D}_{tr}^u|} \frac{1}{HW} \sum_{x_i \in \set{D}_{tr}} \sum_{p\in x_i} \mathcal{L}_{st}(p), \\
\end{aligned}
\end{equation}
where $\mathcal{L}_{st}(p)$ denotes the point-wise self-training loss, $H$ and $W$ are the height and width of the image.

\noindent \textbf{Semantic-Aware Hierarchical Normalization:}  
Depth normalization is a crucial component in our self-training framework, designed to resolve the inherent scale and shift ambiguity in both the teacher's pseudo-labels $d^*$ and the student's predictions $d$. For each pixel $p$ with depth $d_p$ and associated context $\mathcal{C}_p$, normalization maps the depth into a canonical space using robust statistics:
\begin{equation}\label{eq:norm}
\Phi(d_{p}, \mathcal{C}_{p}) = \frac{d_{p} - t( \mathcal{C}_{p})}{s(\mathcal{C}_{p}) + \epsilon}
\end{equation}
where $t$ and $s$ denote the median and the median absolute deviation (MAD) of depths within $\mathcal{C}_p$.

Traditional normalization approaches differ primarily in how they define $\mathcal{C}_p$. {Global Normalization}~\cite{ranftl2020towards, yang2024depth} uses a single context $\mathcal{C}_p = \mathcal{C}_{global}$ containing all valid pixels. In contrast, {Hierarchical Depth Normalization (HDN)}~\cite{zhang2022hierarchical, he2025distill} constructs a multi-scale hierarchy $\mathcal{C}_p = \{\mathcal{C}_{global}, \mathcal{C}_{l-1}, \mathcal{C}_{l-2}, \dots\}$ based on fixed grids or depth bins. However, these content-agnostic groupings often fragment semantic objects, leading to unstable statistics.

We propose {Semantically-Aware Hierarchical Depth Normalization (SA-HDN)} to address these limitations by defining the hierarchy using semantic instance masks. SA-HDN includes two levels: 
\begin{itemize}
    \item The \textbf{Global Context} $\mathcal{C}_{global}$ provides a scene-level normalization using all valid pixels.
    \item The \textbf{Instance Contexts} $\{\mathcal{C}_{ins}^k\}$ capture fine-grained object-level structure, where each $\mathcal{C}_{ins}^k$ contains pixels in a specific object instance $M_k$.
\end{itemize}

We employ SAM2~\cite{ravisam2} to automatically generate $M_k$ on the fly, allowing flexible domain-adaptive context construction without retraining. For each pixel $p$ assigned to instance $k$, the context is defined as $\mathcal{C}_p = \{\mathcal{C}_{global}, \mathcal{C}_{ins}^k\}$, enabling dual supervision at global and instance levels.
The self-training loss at pixel $p$ is then defined as the average MAE between the normalized pseudo-label and prediction over all associated contexts:
\begin{equation}\label{eq:selftrain_hdn}
\begin{aligned}
\mathcal{L}_{st}(p) &= \frac{1}{|\mathcal{C}_p|} \sum_{c \in \mathcal{C}_p} \left| \textbf{sg}(\Phi(d^*_{p}, c)) - \Phi(d_{p}, c) \right| \\
d^*_p &= f(T_w(x_p); \tilde{\Theta}_t), \quad d_p = f(T_s(x_p); \Theta_t)
\end{aligned}
\end{equation}
where $\textbf{sg}(\cdot)$ indicates stop-gradient to block teacher backpropagation. $\Phi$ applies the context-dependent normalization to the predicted relative depth.

\subsection{Weakly Supervised Adaptation}

While self-training strategy described above provides consistent performance gains even without labeled data, its effectiveness is hampered by confirmation bias from pseudo-labels, which can reinforce structural errors. Furthermore, it becomes redundant and offers only marginal gains when the baseline model is already proficient. To overcome these issues, we complement the dense self-training with a weakly supervised objective~\cite{chen2016single}, incorporating a minimal and inexpensive set of relative depth annotations such that $\set{D}_{tr}^w \neq \emptyset$, to guide a more  efficient adaptation.

Each weak label $w_j = \{p_{jn}^+, p_{jn}^-, l_{jn}\}$ represents the ordinal relationship between two pixels. The label $l_{jn} \in \{-1, 0, 1\}$ indicates whether $p_{jn}^+$ is farther, equal in depth, or closer than $p_{jn}^-$, respectively. We use a margin ranking loss to enforce these relations:
\begin{equation}\label{eq:weak_loss}
    \mathcal{L}_{weak} = \sum_{(p_{jn}^+, p_{jn}^-, l_{jn}) \in w_j} \ell(\hat{d}_{jn}^+, \hat{d}_{jn}^-, l_{jn})
\end{equation}
\text{where the loss for a single pair is defined as}
\begin{equation}\label{eq:single_pair_weak_loss}
\begin{aligned}
    \ell(\hat{d}_{jn}^+, \hat{d}_{jn}^-, l_{jn}) = & \mathbf{1}(l_{jn} \neq 0) \cdot \max \left( 0, -l_{jn}\Delta d_{jn} + \delta \right) \\
    &+ \mathbf{1}(l_{jn} = 0) \cdot \left| \Delta d_{jn} \right| \\
    \text{s.t.} \quad \Delta d_{jn} = \hat{d}_{jn}^+ &- \hat{d}_{jn}^- ,\quad d_{jn}^\pm = f(T_w(x_j); \Theta_t)_{p_{jn}^\pm}
\end{aligned}
\end{equation}

\noindent where $\delta$ is a slack variable controlling the margin. This loss encourages depth estimates that are consistent with the weak supervision, even with sparse annotations. Our experiments demonstrate that a small amount of such supervision significantly enhances training stability and generalization.

\subsection{Robust Adaptation with LoRA}

Full fine-tuning of large-scale MDE foundation models is often impractical due to excessive computational demands and limited batch sizes, which lead to inefficiencies and potential overfitting on small target datasets. Additionally, full fine-tuning 
risks catastrophic forgetting of pre-trained knowledge, leading to potentially poorer performance on the target domain.
To address this challenge, we adopt a low-rank adaptation (LoRA) strategy~\cite{hu2022lora}, injecting trainable low-rank matrices into each attention layer $\Theta_a \in \mathbb{R}^{d_1 \times d_2}$ of the encoder. The adapted weights are formulated as:
\begin{equation}
\Theta_a + UV
\end{equation}
where $U \in \mathbb{R}^{d_1 \times r}$ and $V \in \mathbb{R}^{r \times d_2}$ with $r \ll \min(d_1,d_2)$. Only $U$ and $V$ are updated during adaptation, substantially reducing memory and computational overhead. This lightweight tuning mitigates overfitting and catastrophic forgetting while retaining the model’s generalization capabilities.

\noindent \textbf{Weight Regularization:}  
Despite LoRA’s benefits, adaptation under severe domain shifts or noisy pseudo-labels remains susceptible to confirmation bias, where erroneous pseudo-labels may reinforce model errors. To stabilize adaptation, we introduce a weight regularization term that constrains the student weights to remain close to the pre-trained initialization.

Let $U_0, V_0$ be the initial LoRA weights (typically $U_0V_0 = 0$), and $U_t, V_t$ be the student’s current weights. We define a regularization loss that penalizes large deviations:
\begin{equation}\label{eq:reg}
    \mathcal{L}_{reg} = \sum_{U_{tk} \in U_t, V_{tk} \in V_t} \left\| \frac{\alpha}{r} U_{tk} V_{tk} \right\|_2^2
\end{equation}
where $\alpha$ controls the regularization strength. This encourages parameter updates only when strongly justified by new evidence from the target domain, helping to mitigate overfitting and stabilize adaptation.

\subsection{Overall Loss}

Starting from a public depth foundation model, we perform adaptation using a self-training framework enhanced with weight regularization and weak supervision. The total loss is a weighted sum of three components.

\begin{equation}
    \mathcal{L} = \lambda_{st}\mathcal{L}_{st} + \lambda_w \mathcal{L}_{weak} + \lambda_r \mathcal{L}_{reg}
\end{equation}

\section{Experiment}

\subsection{Datasets and Metrics}
We evaluate our method WeSTAR on several widely-used datasets across diverse scenarios. These datasets exhibit significant domain shifts from the training data, providing a robust testbed for assessing the generalization ability of our method under diverse distribution shifts. The evaluated datasets include: \textbf{NYU-V2}~\cite{silberman2012indoor}, \textbf{KITTI}~\cite{geiger2013vision}, \textbf{Sintel}~\cite{butler2012naturalistic}, \textbf{DIODE}~\cite{vasiljevic2019diode}, and their correspoding corrupted benchmarks \textbf{NYU-C}, \textbf{KITTI-C}, \textbf{DIODE-C}, and \textbf{Sintel-C}, following~\cite{kong2023robodepth}. These benchmarks introduce 6 types of corruptions at the highest severity 5 to the original datasets. Moreover, to evaluate our method on more realistic datasets under distribution shift, we choose the night version of \textbf{NuScenes}~\cite{caesar2020nuscenes} and the \textbf{DrivingStrero}~\cite{yang2019drivingstereo} dataset with different weathers~(Sunny, Cloudy, Foggy and Rainy).
More dataset details are provided in the Appendix.

For each dataset, we construct non-overlapping training and test splits. Adaptation is performed only on the training split, while evaluation is carried out on the unseen test set.

\noindent\textbf{Evaluation Metrics}:  Following \cite{yang2024depth, yang2024depthv2}, we adopt two standard metrics to evaluate depth estimation performance on all datasets: $\delta_1$ (the percentage of pixels satisfying $\max (d_i^*/d_i,d_i/d_i^*)<1.25$) and AbsRel (the mean absolute relative error $1/N\sum_{i=1}^N |d_i^*-d_i|/d_i$).

\subsection{Experimental Settings}

\begin{figure*}[!t]
    \centering
    \includegraphics[width=\linewidth]{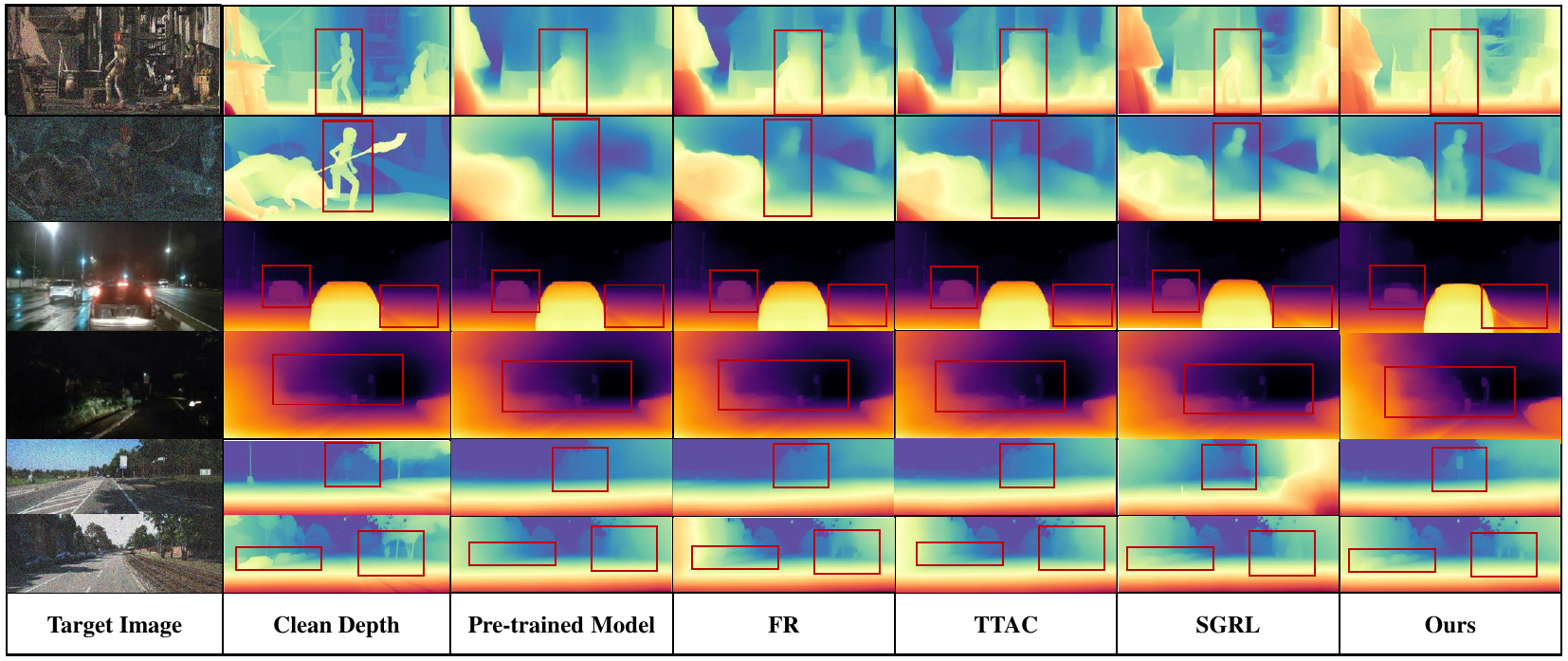}
    \caption{Qualitative results on some selected examples.}
    \vspace{-0.2cm}
    \label{fig:qualitative}
\end{figure*}

\noindent\textbf{MDE Model}: We focus on Relative Depth Estimation and adopt two Relative MDE foundation models, Depth Anything v2~\cite{yang2024depthv2} and MiDas v3.1\cite{birkl2023midas} as our base model, known for their strong zero-shot generalization.

\noindent\textbf{Hyper-parameters:} We use AdamW optimizer with weight decay of 0.0001. For Depth Anything v2 backbone, loss weights $\{\lambda_{st},\lambda_w,\lambda_r\}$ are set to $\{1.0,0.001,1.0\}$. Initial learning rate is set to $0.1$ and scaled linearly with batch size~\cite{goyal2017accurate}, i.e., $0.1 \cdot (\text{BS} / 256)$. A cosine annealing schedule is applied to decay learning rate. We train for up to 100 epochs with early stopping after 30 epochs. The EMA decay factor is 0.996. We use a LoRA with the rank of 8 and alpha of 16. Hyper-Parameters for Midas v3.1 backbone follow a similar configuration, as detailed in the appendix. All experiments are conducted on single NVIDIA RTX 3090 GPU with a batch size of 4.

\noindent \textbf{Pixel-Pairs Sampling:} To construct the ranking loss in weakly-supervised component, we employ a structured pixel-pair sampling strategy instead of sampling pairs independently. For each image, we perform 5 sampling iterations. In the k-th iteration, we first randomly select an anchor pixel $p_k$, then randomly sample a farther point $p_k^+$ and a nearer point $p_k^-$. This results in two structured pairs, $(p_k^+,p_k)$ and $(p_k,p_k^-)$ for supervision. Unlike random sampling, this approach enforces the geometric property of transitivity, providing more coherent and consistent supervision.

\noindent\textbf{Competing Methods}: We compare our method WeSTAR against approaches from source-free domain adaptation and weakly supervised depth estimation. As a baseline, we include the zero-shot performance of the original pretrained model without fine-tuning (\textbf{Source}). Among adaptation methods, \textbf{TTT++}\cite{liu2021ttt++} applies contrastive learning and online feature alignment for improved test-time robustness. \textbf{FR}\cite{eastwood2022sourcefree} uses softly-binned histograms for feature alignment, while \textbf{TTAC}\cite{su2022revisiting} aligns Gaussian feature distributions across domains. \textbf{SSA}\cite{adachi2025test} proposes a regression-specific strategy by aligning features in a dominant subspace. Note that FR, TTAC, and SSA rely on source-domain statistics, which are unavailable in true zero-shot settings, but we include them here for comparison.
We also consider recent ViT-based self-supervised learning methods, which have proven effective for unsupervised feature representation. In particular, we use iBOT\cite{zhou2021ibot} as a representative baseline due to its strong performance. We further fine-tune our encoder module using iBOT and denote this variant as \textbf{iBOT*}. For weak supervision, \textbf{SGRL}\cite{xian2020structure} introduces a structure-guided relative ranking loss. To ensure fairness, all methods are evaluated using the same backbone.

\begin{table*}[htbp]
  \centering
  \small
  \setlength{\tabcolsep}{2pt}{
  \begin{tabular}{l@{\hspace{3mm}}cccccccccccccccccc}
        \toprule
        \multirow{2}{*}{Method} & \multicolumn{2}{c}{NYU} & \multicolumn{2}{c}{KITTI} & \multicolumn{2}{c}{Sintel} & \multicolumn{2}{c}{DIODE} & \multicolumn{2}{c}{NuScenes} & \multicolumn{2}{c}{D-Sunny} & \multicolumn{2}{c}{D-Foggy} & \multicolumn{2}{c}{D-Cloudy} & \multicolumn{2}{c}{D-Rainy} \\
        \cmidrule(lr){2-3} \cmidrule(lr){4-5} \cmidrule(lr){6-7} \cmidrule(lr){8-9} \cmidrule(lr){10-11} \cmidrule(lr){12-13} \cmidrule(lr){14-15} \cmidrule(lr){16-17} \cmidrule(lr){18-19}
        & $\delta_1$ & AbsRel & $\delta_1$ & AbsRel & $\delta_1$ & AbsRel & $\delta_1$ & AbsRel & $\delta_1$ & AbsRel & $\delta_1$ & AbsRel & $\delta_1$ & AbsRel & $\delta_1$ & AbsRel & $\delta_1$ & AbsRel \\
        \midrule
Source &  97.7 & 4.6 & 93.4 & 8.4 & 74.8 & 20.3 & 95.0 & 7.0 & 74.4 & 18.5 & 78.4 & 15.7 & 91.3 & 9.0 & 81.5 & 14.5 & 84.8 & 12.1 \\
TTT++ & 97.7 & 4.6 & 93.4 & 8.3 & 74.8 & 20.2 & \textbf{95.2} & 7.0 & 74.5 & 18.4 & 78.4 & 15.7 & 91.3 & 9.0 & 81.5 & 14.4 & 84.9 & 12.1 \\
TTAC & 97.7 & 4.6 & 93.4 & 8.5 & 75.0 & 20.5 & 95.0 & 7.0 & 74.4 & 18.5 & 79.5 & 15.4 & 91.1 & 9.1 & 82.0 & 14.4 & 84.5 & 12.2\\
FR & 97.6 & 4.7 & 93.8 & 8.2 & 74.6 & 20.6 & 95.0 & 7.4 & 74.4 & 18.5 & 79.1 & 15.5 & 91.3 & 9.0 & 82.3 & 14.3 & 84.8 & 12.1 \\
SSA & 97.8 & 4.5 & 93.5 & 8.4 & 75.5 & 18.9 & 95.1 & 6.9 & 74.3 & 18.4 & 79.6 & 15.3 & 90.8 & 9.1 & 82.1 & 14.4 & 84.7 & 11.9 \\
iBOT* & 97.7 & 4.6 & 93.2 & 8.7 & 74.1 & 20.5 & 95.0 & 7.1 & 74.5 & 18.4 & 78.4 & 15.6 & 91.5 & 8.9 & 81.6 & 14.4 & 84.7 & 12.1\\
SGRL & 97.6 & 4.8 & 94.1 & 7.7 & 76.9 & 21.9 & 95.1 & 6.9 & 75.8 & 17.6 & \textbf{83.4} & \textbf{13.5} & 92.0 & 8.5 & 84.4 & 12.9 & 85.3 & 12.0\\
WeSTAR & \textbf{98.2} & \textbf{4.3} & \textbf{95.1} & \textbf{7.2} & \textbf{82.2} & \textbf{16.9} & \textbf{95.2} & \textbf{6.5} & \textbf{78.1} & \textbf{16.2} & 82.8 & 13.6 & \textbf{92.5} & \textbf{8.2} & \textbf{85.4} & \textbf{12.5} & \textbf{87.4} & \textbf{10.6} \\

        \bottomrule
    \end{tabular}
    }
    \caption{Adaptation results on unseen realistic datasets. For $\delta_1$, higher is better, while for AbsRel, lower is better.}
    \label{clean_datasets}
    \vspace{-0.3cm}
\end{table*}

\subsection{Adaptation Results on Diverse Benchmarks}
\noindent \textbf{Adaptation to Corrupted Datasets:}
We present the comparison results on four corrupted benchmarks and report the average performance over all 6 corruptions in Table~\ref{tab:unseen_avg} (detailed results on each corruption are deferred to the Appendix).
When the target images are affected by data corruptions, sensor failures, or adverse weather conditions, the performance of the base model degrades significantly. 
Specifically, the average performance of NYU-C across 6 corruption types drops from $\delta_1$: 97.7 to 87.4 (a 10.5\% decrease) and AbsRel: 4.6 to 10.6 (a 56.6\% increase). These results underscore the substantial challenge that domain shifts pose to the zero-shot generalization capability of monocular depth estimation (MDE) foundation models.

When SFDA methods such as TTAC, FR, and SSA are applied, they consistently improve performance compared to the baseline model. In contrast, applying TTT++ results in a noticeable degradation in performance in KITTI-C. This suggests that simple contrastive loss and feature alignment mechanisms are insufficient to effectively handle the distribution shifts caused by the corruptions.

Meanwhile, the self-supervised method iBOT* demonstrates competitive performance without access to the ground truth, highlighting the strength of robust feature learning under severe domain shifts. Similarly, the weakly supervised method SGRL also achieves strong results, illustrating the potential of weak supervision for improving robustness in the presence of corruptions.
Notably, WeSTAR consistently achieves state-of-the-art performance across nearly all corruption types, clearly demonstrating superior generalization under challenging distribution shifts. 

We observe a similar results on other datasets, but TTT++ shows better performance than the unadapted baseline model. WeSTAR maintains a consistent advantage across the majority of corruption types, particularly under severe distortions. The performance gains are especially pronounced on the more challenging synthetic scenes in the Sintel-C dataset, further validating the robustness and generalization capability of our WeSTAR.
Notably, with the MiDaS backbone, TTT++ outperforms other SFDA methods, while WeSTAR maintains a clear advantage, further validating its effectiveness in various backbones. We excluded NYU-C and KITTI-C since their clean versions are part of MiDaS' training data, ensuring fair evaluation.

\begin{table}[htbp]
    \centering
    \small
    \setlength{\tabcolsep}{1.5pt}
    {
    \begin{tabular}{l@{\hspace{3mm}}cccccccc}
    \toprule
    \multirow{2}{*}{Method} & \multicolumn{2}{c}{NYU-C} & \multicolumn{2}{c}{KITTI-C} & \multicolumn{2}{c}{Sintel-C} & \multicolumn{2}{c}{DIODE-C}\\
    \cmidrule(lr){2-3}\cmidrule(lr){4-5}\cmidrule(lr){6-7}\cmidrule(lr){8-9}
     & $\delta_1$ & AbsRel & $\delta_1$ & AbsRel & $\delta_1$ & AbsRel & $\delta_1$ & AbsRel\\
    \midrule
    Source & 87.4 & 10.6 & 83.2 & 13.2 & 60.3 & 30.6 & 88.0 & 11.1\\
    TTT++  & 90.1 & 9.7 & 70.2 & 19.4 & 62.7 & 27.1 & 88.7 & 10.5\\
    TTAC   & 91.2 & 8.9 & 84.7 & 12.7 & 62.3 & 29.7 & 89.2 & 10.6\\
    FR     & 91.2 & 8.8 & 84.0 & 12.9 & 62.6 & 29.5 & 89.2 &10.5\\
    SSA    & 91.3 & 8.8 & 84.2 & 12.7 & 63.7 & 28.5 & 88.9 & 10.6\\
    iBOT*   & 92.1 & 8.5 & 85.6 & 12.5 & 62.7 & 29.0 & 89.6 & 10.3\\
    SGRL   & 92.4 & 8.4 & 87.4 & 11.3 & 66.5 & 29.9 & 90.0 & 10.0 \\
    WeSTAR   & \textbf{94.6} & \textbf{7.1} & \textbf{88.7} & \textbf{10.5} & \textbf{71.8} & \textbf{24.1} & \textbf{91.3} & \textbf{9.3} \\

    \bottomrule
    \end{tabular}
    }
    \caption{Adaptation results averaged over all 6 types of corruptions on corrupted datasets.}
    \label{tab:unseen_avg}
\end{table}

\noindent \textbf{Results on Realistic Datasets With Diverse Scenarios:}
As shown in Table~\ref{clean_datasets}, the powerful Depth Anything V2 model already exhibits a high performance ceiling on standard benchmarks such as NYU, KITTI, and DIODE, leaving limited room for further improvements, where most adaptation methods struggle to yield further gains and may even degrade performance. In contrast, our proposed WeSTAR consistently  maintains or improves upon this strong baseline. Notably, the benefits of WeSTAR are especially evident on the synthetic Sintel dataset and in realistic scenarios involving complex environmental changes, such as varying weather and lighting conditions. This highlights the superior generalization ability of our WeSTAR, even when adapting to clean yet distributionally distinct target domains.
We also reach similar conclusions on Midas v3.1 backbone in Table~\ref{midas_results}. To ensure fair evaluation, we excluded NYU and KITTI as they are included in the training data.

\begin{table*}[htbp]
  \centering
  \small
  \setlength{\tabcolsep}{2pt}{
  \begin{tabular}{l@{\hspace{3mm}}cccccccccccccccccc}
        \toprule
        \multirow{2}{*}{Method} & \multicolumn{2}{c}{Sintel-C} & \multicolumn{2}{c}{DIODE-C} & \multicolumn{2}{c}{Sintel} & \multicolumn{2}{c}{DIODE} & \multicolumn{2}{c}{NuScenes} & \multicolumn{2}{c}{D-Sunny} & \multicolumn{2}{c}{D-Foggy} & \multicolumn{2}{c}{D-Cloudy} & \multicolumn{2}{c}{D-Rainy} \\
        \cmidrule(lr){2-3} \cmidrule(lr){4-5} \cmidrule(lr){6-7} \cmidrule(lr){8-9} \cmidrule(lr){10-11} \cmidrule(lr){12-13} \cmidrule(lr){14-15} \cmidrule(lr){16-17} \cmidrule(lr){18-19}
        & $\delta_1$ & AbsRel & $\delta_1$ & AbsRel & $\delta_1$ & AbsRel & $\delta_1$ & AbsRel & $\delta_1$ & AbsRel & $\delta_1$ & AbsRel & $\delta_1$ & AbsRel & $\delta_1$ & AbsRel & $\delta_1$ & AbsRel \\
        \midrule
Source & 43.2 & 40.5 & 74.6 & 17.7 & 58.6 & 30.4 & 91.4 & 9.3 & 52.3 & 32.5 & 73.7 & 17.8 & 84.2 & 11.9 & 77.2 & 16.4 & 62.2 & 22.4 \\
TTT++ & 48.4 & 36.6 & 81.7 & 14.3 & 61.5 & 26.6 & 91.4 & 9.3 & 53.0 & 32.3 & 75.9 & 16.8 & 84.5 & 11.9 & 78.1 & 15.9 & 67.5 & 18.6 \\
TTAC & 44.8 & 38.3 & 80.2 & 15.0 & 58.7 & 30.4 & 91.4 & 9.3 & 52.3 & 32.5 & 73.6 & 17.8 & 84.2 & 11.9 & 77.2 & 16.4 & 62.1 & 22.5 \\
FR & 45.2 & 38.0 & 81.4 & 14.7 & 58.6 & 30.4 & 91.4 & 9.3 & 53.6 & 32.0 & 73.5 & 17.9 & 84.5 & 11.7 & 78.5 & 15.7 & 69.4 & 18.6 \\
SSA & 45.5 & 38.8 & 80.3 & 15.2 & 58.6 & 30.4 & 91.4 & 9.3 & 53.1 & 31.8 & 75.2 & 17.2 & 83.7 & 12.0 & 78.1 & 15.8 & 70.7 & 18.0 \\
SGRL & 52.0 & 35.6 & 85.9 & 12.2 & 65.5 & 26.2 & 91.6 & 9.1 & 63.4 & 24.4 & \textbf{79.8} & \textbf{15.2} & 85.3 & 11.2 & 79.1 & 15.0 & 74.5 & 16.2 \\
WeSTAR & \textbf{56.6} & \textbf{34.8} & \textbf{86.6} & \textbf{12.0} & \textbf{67.9} & \textbf{25.8} & \textbf{91.7} & \textbf{9.0} & \textbf{67.5} & \textbf{22.0} & 78.3 & 15.8 & \textbf{87.1} & \textbf{10.1} & \textbf{80.4} & \textbf{14.5} & \textbf{75.3} & \textbf{15.5} \\

        \bottomrule
    \end{tabular}
    }
    \caption{Adaptation results for MiDas v3.1 backbone.}
    \label{midas_results}
    \vspace{-0.3cm}
\end{table*}

\noindent \textbf{Qualitative Studies:}
We present qualitative results on corrupted and realistic test samples in Figure~\ref{fig:qualitative}. The comparison includes predictions on the corresponding clean images (Clean Depth), predictions from the pre-trained model without adaptation (Pre-trained Model), and results from various adaptation methods. Note that when the target image is not the corrupted version, "clean depth" and "Pretrained Model" is the same result. Overall, our proposed method WeSTAR consistently produces more accurate depth estimates, often surpassing even the predictions on clean images. More visualization results are provided in appendix.

\subsection{Ablation Study and Additional Analysis}
\noindent\textbf{Effects of Individual Components}:
We conduct ablation studies on three datasets as shown in Table~\ref{ablation_main}.
Starting from the zero-shot baseline, applying self-training (ST) consistently improves performance, especially on the corrupted Sintel-C dataset~($\delta_1$: 60.3 $\to$ 63.3, AbsRel: 30.6 $\to$ 27.6), indicating enhanced robustness under domain shifts. However, ST brings only marginal gains on clean datasets like Sintel and NuScenes, due to confirmation bias from inaccurate pseudo-labels and the strong geometric prior of the pre-trained model, which limits further improvement from ST alone.
Weak supervision~(WS) moderately improves $\delta_1$ (e.g., 74.8 $\to$ 77.5 on Sintel), but its scale insensitivity introduces geometric distortions and yields inconsistent AbsRel results, sometimes even degrading performance (e.g., 20.3 $\to$ 24.1 on Sintel).
Weight regularization~(WR) further stabilizes training by anchoring to initial parameters, reducing overfitting to noisy pseudo-labels or ordinal relations.
Combining all components, WeSTAR significantly outperforms all variants, highlighting their complementary effects.

\begin{table}[htbp]
  \centering
  \small
  \setlength{\tabcolsep}{4pt}
  \begin{tabular}{ccccccccc}
        \toprule
        \multirow{2}{*}{ST} & \multirow{2}{*}{\makecell[c]{WR}} & \multirow{2}{*}{\makecell[c]{WS}} & \multicolumn{2}{c}{Sintel-C} & \multicolumn{2}{c}{Sintel} & \multicolumn{2}{c}{NuScenes} \\
        \cmidrule(lr){4-5} \cmidrule(lr){6-7} \cmidrule(lr){8-9}
        & & & $\delta_1$ & AbsRel & $\delta_1$ & AbsRel & $\delta_1$ & AbsRel \\
        \midrule
        - & - & - & 60.3 & 30.6 & 74.8 & 20.3 & 74.4 & 18.5 \\
        \checkmark & - & - & 63.3 & 27.6 & 74.9 & 20.3 & 74.4 & 18.4  \\
        - & - & \checkmark & 68.3 & 28.2 & 77.5 & 24.1 & 76.9 & 16.9 \\
        \checkmark & - &\checkmark & 70.1 & 27.1 & 80.2 & 21.3 & 76.8 & 16.7 \\
       \checkmark & \checkmark &\checkmark & \textbf{71.8} & \textbf{24.1} & \textbf{82.2} & \textbf{16.9} & \textbf{78.1} & \textbf{16.2} \\
        
        \bottomrule
    \end{tabular}
    \caption{Ablation Studies of Individual Component. "ST" is the self-training loss, "WS" is the weakly-supervised loss and "WR" is the weight regularization loss.}
    \label{ablation_main}
    \vspace{-0.3cm}
\end{table}

\noindent\textbf{Effects of Tuning Different Modules}: 
Table~\ref{ablation_update_encoder} shows that LoRA-based adaptation consistently outperforms full fine-tuning. Although full fine-tuning slightly outperforms LoRA on Sintel-C, this likely stems from its tendency to overfit corruptions. In contrast, LoRA constrains updates to a low-rank subspace, mitigating overfitting and preserving pretrained knowledge. Tuning only the decoder brings no improvement, and updating both encoder and decoder offers no advantage over encoder-only tuning, emphasizing the encoder’s key role in adaptation.

\begin{table}[htbp]
  \centering
  \small
  \setlength{\tabcolsep}{4pt}
  \begin{tabular}{lccccccc}
        \toprule
        \multicolumn{1}{r}{\multirow{2}{*}{Component}} & \multicolumn{2}{c}{Sintel-C} & \multicolumn{2}{c}{Sintel} & \multicolumn{2}{c}{NuScenes} \\
        \cmidrule(lr){2-3} \cmidrule(lr){4-5} \cmidrule(lr){6-7}
        & $\delta_1$ & AbsRel & $\delta_1$ & AbsRel & $\delta_1$ & AbsRel  \\
        \midrule
        All Params & \textbf{72.4} & 24.8 & 80.8 & 19.5 & 77.1 & 16.4 \\
        Encoder & 72.2 & 24.7 & 80.7 & 19.4 & 77.2 & 16.6  \\
        Decoder & 60.3 & 30.7 & 74.8 & 20.5 & 74.7 & 18.3 \\
        
        LoRA & 71.8 & \textbf{24.1} & \textbf{82.2} & \textbf{16.9} & \textbf{78.1} & \textbf{16.2} \\ 
        \bottomrule
    \end{tabular}
    \caption{Effects of tuning different modules.}
  \label{ablation_update_encoder}
\end{table}

\noindent\textbf{Effects of Normalization:}
Table~\ref{ablation_norm} validates SA-HDN against Global Norm and HDN. Our method clearly outperforms both, consistently surpassing the strong Global Norm baseline and the grid-based HDN. This is due to its ability to enforce geometric consistency at multiple semantic scales. By normalizing within instances, it preserves fine-grained structure often lost in global alignment.

\begin{table}[htbp]
  \centering
  \small
  \setlength{\tabcolsep}{4pt}
  \begin{tabular}{lccccccc}
        \toprule
        \multicolumn{1}{r}{\multirow{2}{*}{Norm}} & \multicolumn{2}{c}{Sintel-C} & \multicolumn{2}{c}{Sintel} & \multicolumn{2}{c}{NuScenes} \\
        \cmidrule(lr){2-3} \cmidrule(lr){4-5} \cmidrule(lr){6-7}
        & $\delta_1$ & AbsRel & $\delta_1$ & AbsRel & $\delta_1$ & AbsRel  \\
        \midrule
        Global & 70.9 & 24.2 & 80.8 & \textbf{16.4} & 77.2 & 16.7 \\
        HDN &  68.9 & 28.1 & 78.6 & 22.4 & 77.2 & 16.6 \\
        SA-HDN & \textbf{71.8} & \textbf{24.1} & \textbf{82.2} & 16.9 & \textbf{78.1} & \textbf{16.2} \\ 
        \bottomrule
    \end{tabular}
    \caption{Effects of different Normalization methods.}
  \label{ablation_norm}
\end{table}

\begin{figure}
    \centering
    \includegraphics[width=1.0\linewidth]{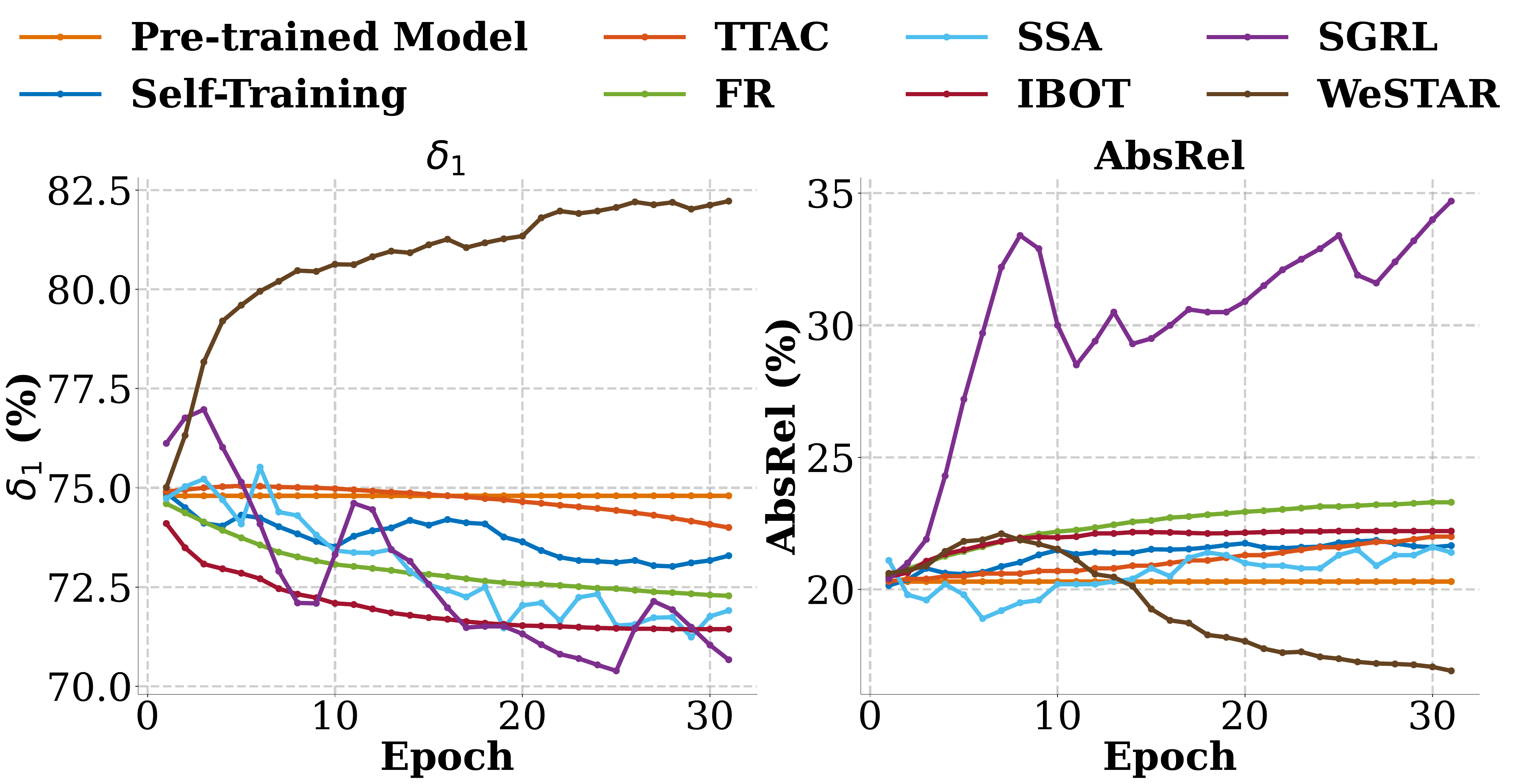}
    \vspace{-0.3cm}
    \caption{Comparative Performance over Training Epochs on Sintel.}
    \vspace{-0.4cm}
    \label{fig:sintel_diode_epochs}
\end{figure}

\noindent\textbf{Training Stability}:  
We evaluate training stability by examining performance trajectories on Sintel in Figure~\ref{fig:sintel_diode_epochs}. Feature alignment and contrastive methods (e.g., TTAC, FR) show modest, inconsistent gains due to sensitivity to low-quality pseudo-labels and feature drift.
SSA yields smoother but conservative improvements, avoiding abrupt changes yet limiting progress. The self-supervised method iBOT achieves rapid early gains but quickly plateaus, suggesting pretrained features alone are insufficient for sustained adaptation under corruption.
SGRL, a weakly supervised approach, improves sharply initially but degrades over time, especially on DIODE Indoor-C, due to limited and noisy supervision causing divergence toward suboptimal solutions.
By contrast, WeSTAR delivers stable, consistent improvements on both $\delta_1$ and AbsRel through training, benefiting from reliable task-specific supervision and effective regularization that prevent performance drops and model drift.

\section{Conclusion}

This work investigates the generalization capabilities of pre-trained depth estimation foundation models under unseen and corrupted conditions. When limited unlabeled and weakly labeled target domain data is available, we propose an adaptation strategy that combines self-training with weight regularization and optional ordinal supervision. This design effectively addresses challenges posed by unreliable pseudo-labels and facilitates stable adaptation. Our comprehensive experiments across diverse benchmark datasets validate the efficacy of our method, showing substantial improvements in both realistic and challenging corrupted domains. The results underscore the potential of combining weak supervision and regularized adaptation to enhance the robustness and applicability of depth estimation foundation models in real-world scenarios.

\section{Acknowledgments}
This work is supported by the Guangdong Basic and Applied Basic Research Foundation (No. 2023A1515110077) and the Agency for Science, Technology and Research (A*STAR) under its MTC Programmatic Funds (Grant No. M23L7b0021).

\bibliography{aaai2026}

\end{document}